\documentclass[conference]{IEEEtran}
\IEEEoverridecommandlockouts
\usepackage{cite}
\usepackage{amsmath,amssymb,amsfonts}
\usepackage{algorithmic}
\usepackage{graphicx}
\usepackage{textcomp}
\usepackage{xcolor}
\usepackage{booktabs}
\usepackage{longtable}
\usepackage{multirow}
\usepackage[backref]{hyperref}

\def\BibTeX{{\rm B\kern-.05em{\sc i\kern-.025em b}\kern-.08em
    T\kern-.1667em\lower.7ex\hbox{E}\kern-.125emX}}
\begin{document}

\title{EfficientRep: An Efficient Repvgg-style ConvNets with Hardware-aware Neural Network Design\\
%{\footnotesize \textsuperscript{*}}
}

%\auchor{\IEEEauthorblockN{Kaiheng Weng}}
\author{
    \IEEEauthorblockN{
        Kaiheng Weng, % \IEEEauthorrefmark{1},
        Xiangxiang Chu, % \IEEEauthorrefmark{1},
        Xiaoming Xu, %\IEEEauthorrefmark{3},
        Junshi Huang, and %\IEEEauthorrefmark{4}, and
        Xiaoming Wei %\IEEEauthorrefmark{5}
    }
    \IEEEauthorblockA{Meituan Inc.}
    \IEEEauthorblockA{\{wengkaiheng, chuxiangxiang, xuxiaoming04, huangjunshi,weixiaoming\}@meituan.com}
}

% \author{\IEEEauthorblockN{Kaiheng Weng}
% \IEEEauthorblockA{\textit{Meituan} \\
% Beijing, China \\
% wengkaiheng@meituan.com}
% \and
% \IEEEauthorblockN{Xiangxiang Chu}
% \IEEEauthorblockA{\textit{Meituan} \\
% Beijing, China \\
% \and
% \IEEEauthorblockN{Xiaoming Xu}
% \IEEEauthorblockA{\textit{Meituan} \\
% Beijing, China \\
% \and
% \IEEEauthorblockN{Junshi Huang}
% \IEEEauthorblockA{\textit{Meituan} \\
% Beijing, China \\
% \and
% \IEEEauthorblockN{Xiaoming Wei}
% \IEEEauthorblockA{\textit{Meituan} \\
% Beijing, China \\

\maketitle
\begin{abstract}
We present a hardware-efficient architecture of convolutional neural network,
which has a repvgg-like architecture. Flops or parameters are traditional metrics to 
evaluate the efficiency of networks which are not sensitive to hardware including computing
ability and memory bandwidth. Thus, how to design a neural network to efficiently use 
the computing ability and memory bandwidth of hardware is a critical problem. This paper proposes a 
method how to design hardware-aware neural network. Based on this method, we designed EfficientRep 
series convolutional networks, which are high-computation hardware(e.g. GPU) friendly and applied 
in YOLOv6 object detection framework. YOLOv6 has 
published YOLOv6N/YOLOv6S/YOLOv6M/YOLOv6L models in v1 and v2 versions. 
Our YOLOv6 code is made available at 
\href{https://github.com/meituan/YOLOv6}{https://github.com/meituan/YOLOv6}.
\end{abstract}


\section{Introduction}

Since VGG achieved success in image classification tasks, convolutional neural network design has attracted 
huge attentions in academy and industry. Currently, amounts of classical networks have been proposed, like
Inception\cite{Inception} and Resnet\cite{Resnet}. These well-designed architectures make the accuracy of 
image classification increasingly higher. Besides manully design, recently neural architecture search  
also automatically designed several representative networks, such as Nasnet\cite{Nasnet} and AmoebaNet\cite{AmoebaNet}.
Though complexed networks bring successes to vision tasks like image classification, object detection and segmentation, 
these networks may not get suitable accuracy-speed balance on deployed hardware. 

\begin{figure}[t]
\centerline{\includegraphics[width=0.8\linewidth]{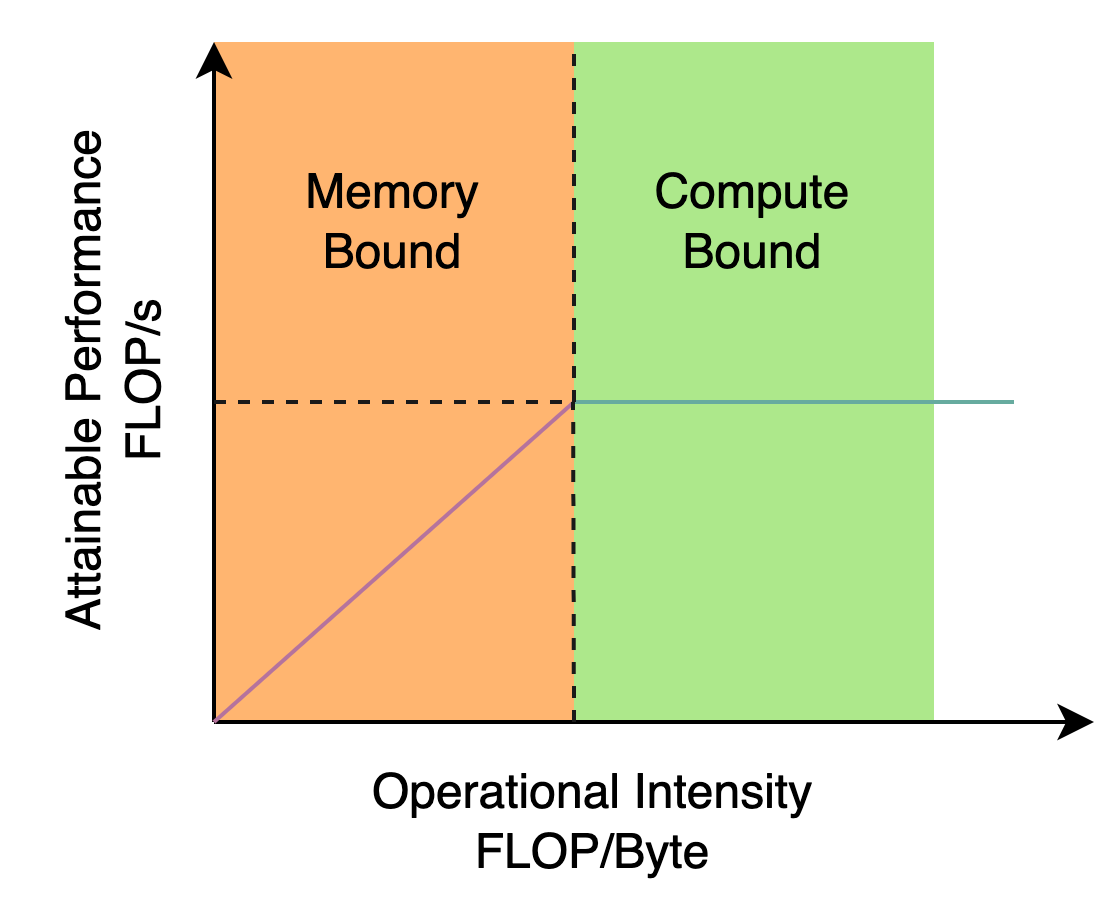}}
\caption{Introduction of Roofline Model.}
\label{Introduction of Roofline Model}
\end{figure}

Deep learning network design and deployment for hardware efficiency has been consistently studied\cite{MobileOne}\cite{TRT-VIT}. Traditional 
evaluation metrics for inference effciency are floating-point operations (FLOPs) and parameter count. However, such metrics 
can not represent the relationships with hardware, such as memory access cost and I/O throughput. Fig \ref{Introduction of Roofline Model} shows 
the relationships between computing ability with memory resources. Hence, an important question
is raised to us: \textit{\textbf{how to design a hardaware-friendly network to perform better accuracy-speed balance?}}

\begin{figure}[t]
\centerline{\includegraphics[width=1.0\linewidth]{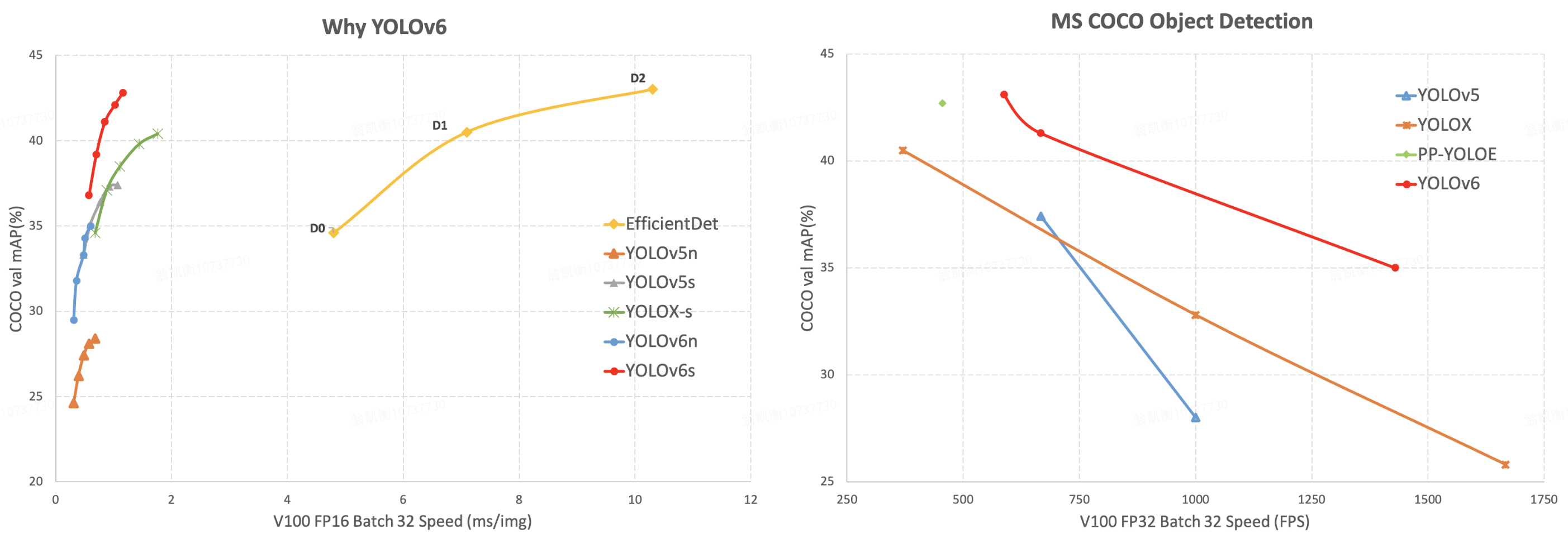}}
\caption{Metrics of YOLOv6-v1.}
\label{Metrics of YOLOv6-v1}
\end{figure}

To solve this problem, we explored some novel architectures that show competitive and applied in YOLOv6 object detection framework.
RepVGG\cite{repvgg} is a novel network with 3x3 convolutional kernel highly optimized by winograd algorithm on GPU or CPU. Single-path 
models can train and infer fast on devices like GPU. Fig \ref{Design of Rep Conv}. shows the transformation between training state and infenrece state for 
rep conv. In training state, with extra 1x1 conv and identity, rep conv can guarantee the accuracy during training. 
In inference state, re-parameterization structure can be equally converted to inference status. 

In YOLOv6-1.0, based on basic rep conv, we designed pure regvgg-style architectures named EfficientRep backbone and Rep-PAN neck to efficiently 
utilize the computing resources of GPU. 
In YOLOv6-2.0, to balance the computing and memory resources, we explored a kind of novel structure named Bep(Beer-mug) unit and BepC3(or named 
CSPStackRep) block. 
Compared with rep conv, we found that Bep unit is a more efficient basic neural network unit at some states. 
Though RepVGG proposed that rep-style multi-branch training 
can reach comparable performance as an original multi-branch training like resnet, we discovered that the accuracy-speed trade-off of rep-style network 
degraded at large scale which will be shown in Section III. Hence, we designed CSPBep backbone and CSPRepPAN neck applied in 
YOLOv6-2.0, with above structures of Bep unit and BepC3 block. 

\begin{figure}[t]
\centerline{\includegraphics[width=1.0\linewidth]{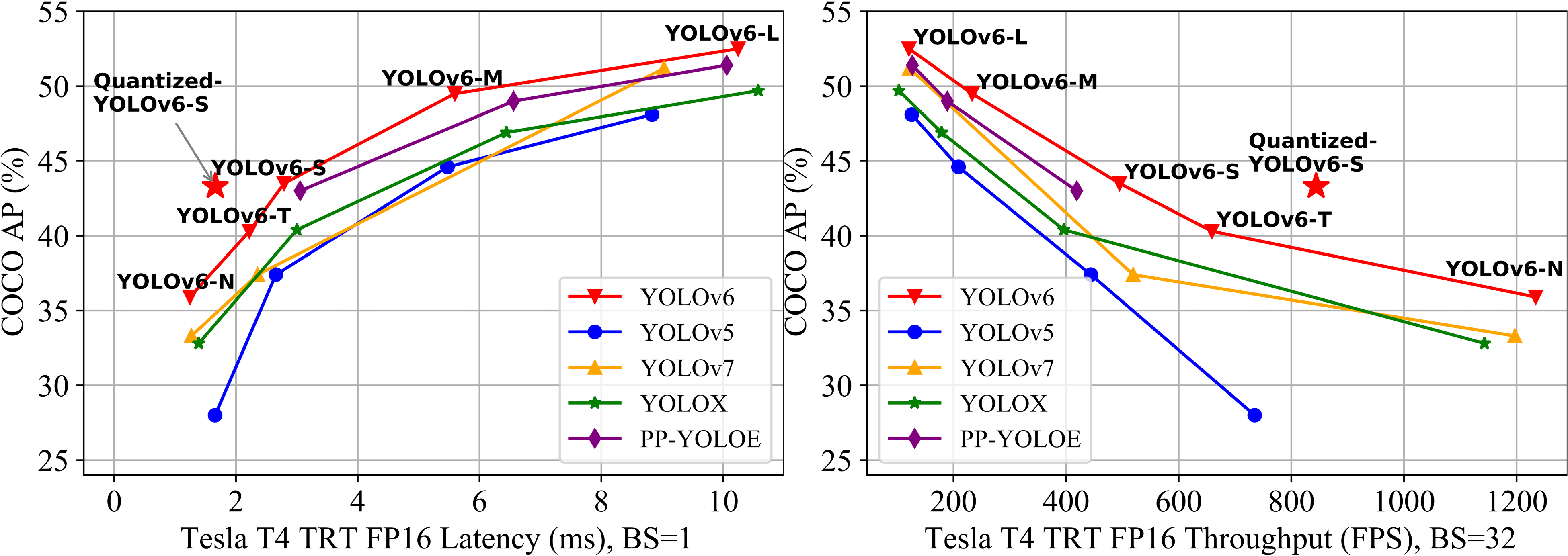}}
\caption{Metrics of YOLOv6-v2.}
\label{Metrics of YOLOv6-v2}
\end{figure}

With above considerations, we apply hybrid strategy to YOLOv6 that we select single-branch models in small size and multi-branch models in 
large size. Fig \ref{Metrics of YOLOv6-v1} and Fig \ref{Metrics of YOLOv6-v2} describe the comparable accuracy-speed trade-off as 
other object detectors. 

\section{Related Work}
\subsection{Neural Network Design}
VGG\cite{VGG} network achieved the top-1 accuracy of ImageNet Classification to above 70\%, many related innovations 
have been proposed, e.g. GoogLeNet\cite{GoogLeNet} and Inception\cite{Inception} network are designed in multi-branch architecture. 
ResNet\cite{Resnet} is a representative two-branch network and widely used in industry. Until repvgg\cite{repvgg} is proposed,
single-path network shows high efficiency on some devices. 

\subsection{Neural Architecture Search}
Neural architecture search (NAS) is a novel technique, aiming to automatically design network compared 
with manual designing. With manual-designing space design, NAS can automatically generate numbers of 
networks with huge resource cost. Currently, to save computing resources, low-cost neural network search has 
been proposed, e.g. One-For-All\cite{one-for-all} et al. 

\subsection{Hardware-aware Neural Network Design}
Recently, serveral novel networks like MobileOne\cite{MobileOne} and TRT-VIT\cite{TRT-VIT} have been proposed. These networks
are manually designed, considering the performance on devices. Except for the metrics about accuracy or params, the inference speed 
is taken into accounts simultaneously, which is called hardware-aware neural network design. 

\subsection*{Our contributions are summarized as follows.}
\begin{itemize}
    \item We proposed novel structures of Bep unit, Repblock and BepC3 block.
    \item We proposed novel network of EfficientRep, Rep-PAN, CspBep and CSPRepPAN.
    \item We proposed novel network design with computing ability and memory bandwidth balance, with various strategies for different size of models.
\end{itemize}

\begin{figure}[t]
\centerline{\includegraphics[width=0.85\linewidth]{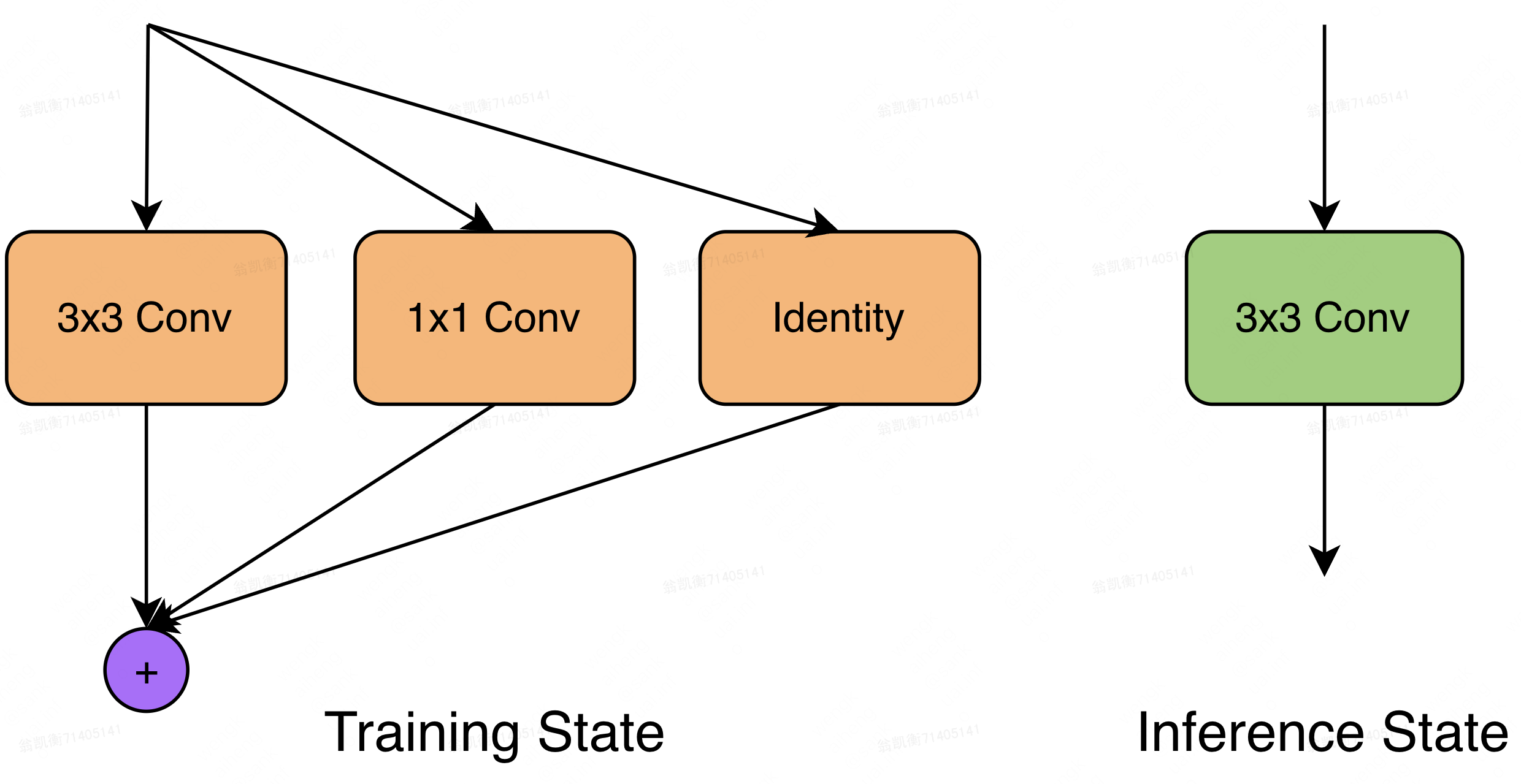}}
\caption{Design of Rep Conv.}
\label{Design of Rep Conv}
\end{figure}

\begin{figure}[b]
\centerline{\includegraphics[width=0.7\linewidth]{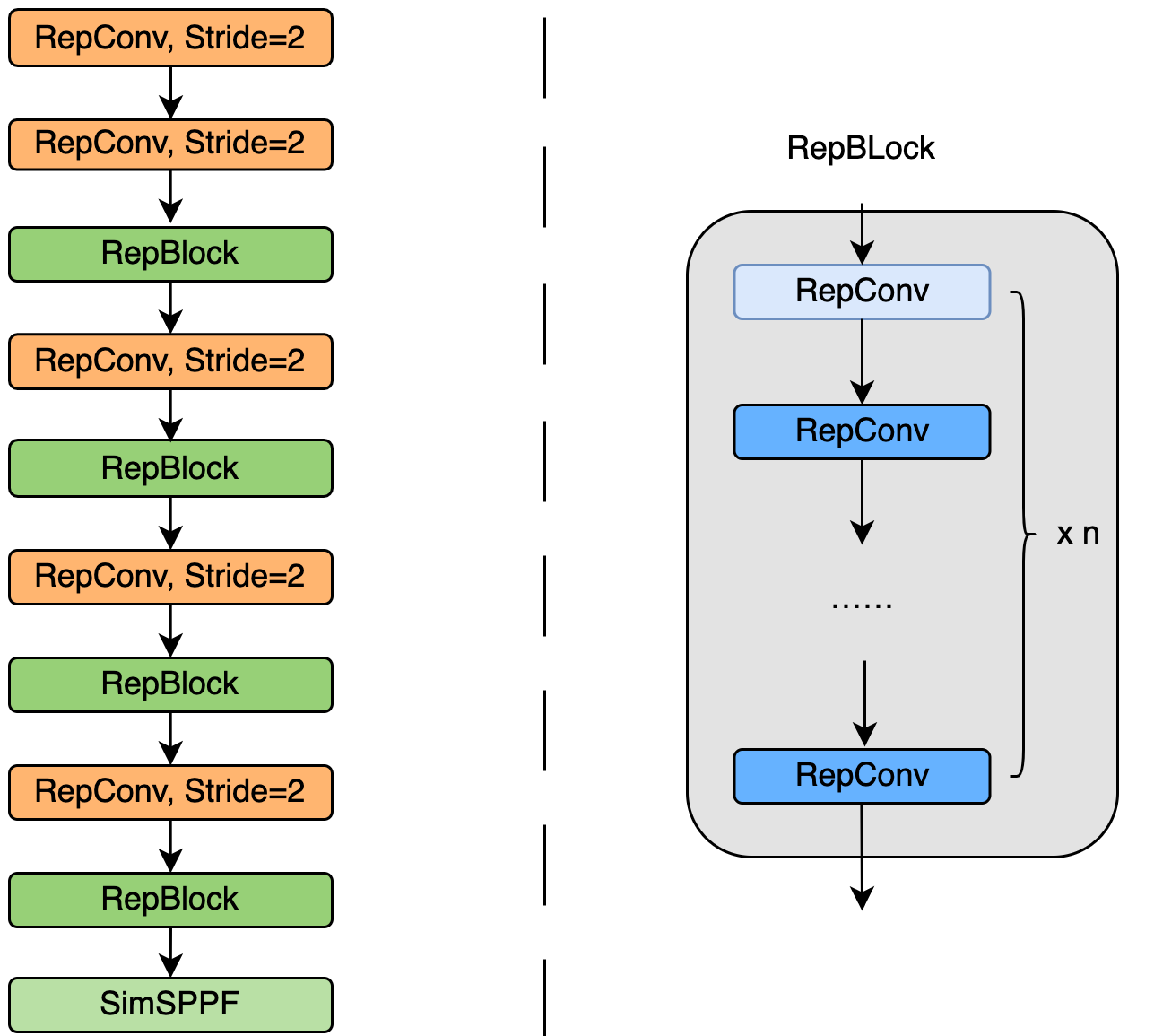}}
\caption{Design of EfficientRep.}
\label{Design of EfficientRep}
\end{figure}

\begin{figure}[t]
\centerline{\includegraphics[width=0.95\linewidth]{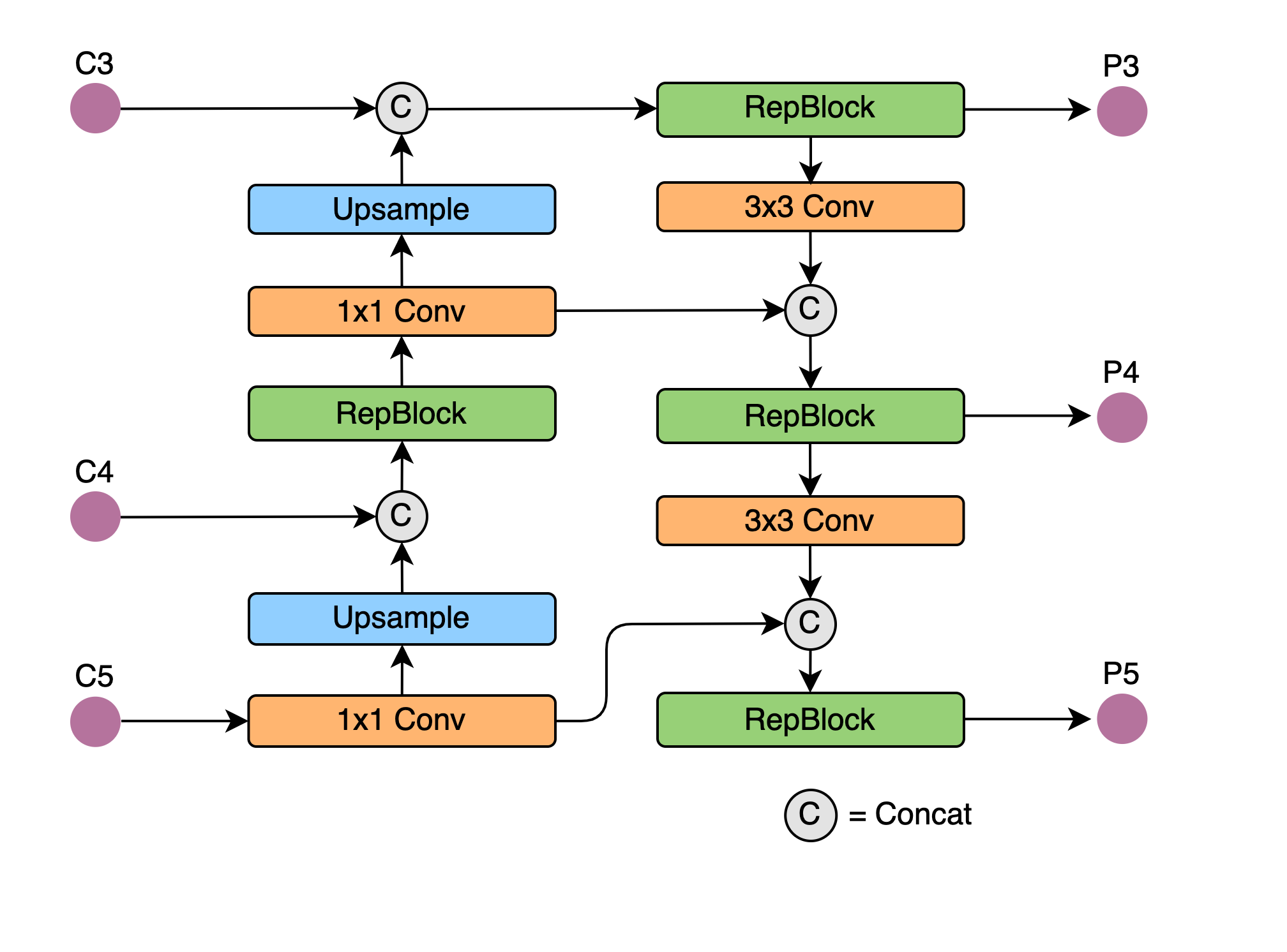}}
\caption{Design of Rep-PAN.}
\label{Design of Rep-PAN}
\end{figure}

\section{Approach}
In this section, we describe the details of hardware-aware neural network design applied in YOLOv6. 
The novel structures and networks we proposed 
will be presented with hybrid strategy for different size of models. 
\subsection{Pure Repvgg-style Efficient Design}
Repvgg-style\cite{repvgg} conv has the structure of 3x3 conv followed by ReLU, which 
can efficiently utilize the hardware computation. At training state, Repvgg-style 
conv consists of 3x3 branch, 1x1 branch and identity(Fig \ref{Design of Rep Conv}). Through re-parameterization,  
multi-branch structure is transformed to single-branch 3x3 conv at inference state. 
As shown in Fig \ref{Design of EfficientRep} and Fig \ref{Design of Rep-PAN}, we designed Repvgg-style network called EfficientRep backbone and Rep-PAN neck which are GPU-friendly, 
and applied in YOLOv6 detection framework(YOLOv6-v1)\cite{YOLOv6}. 

\begin{figure}[b]
\centerline{\includegraphics[width=0.3\linewidth]{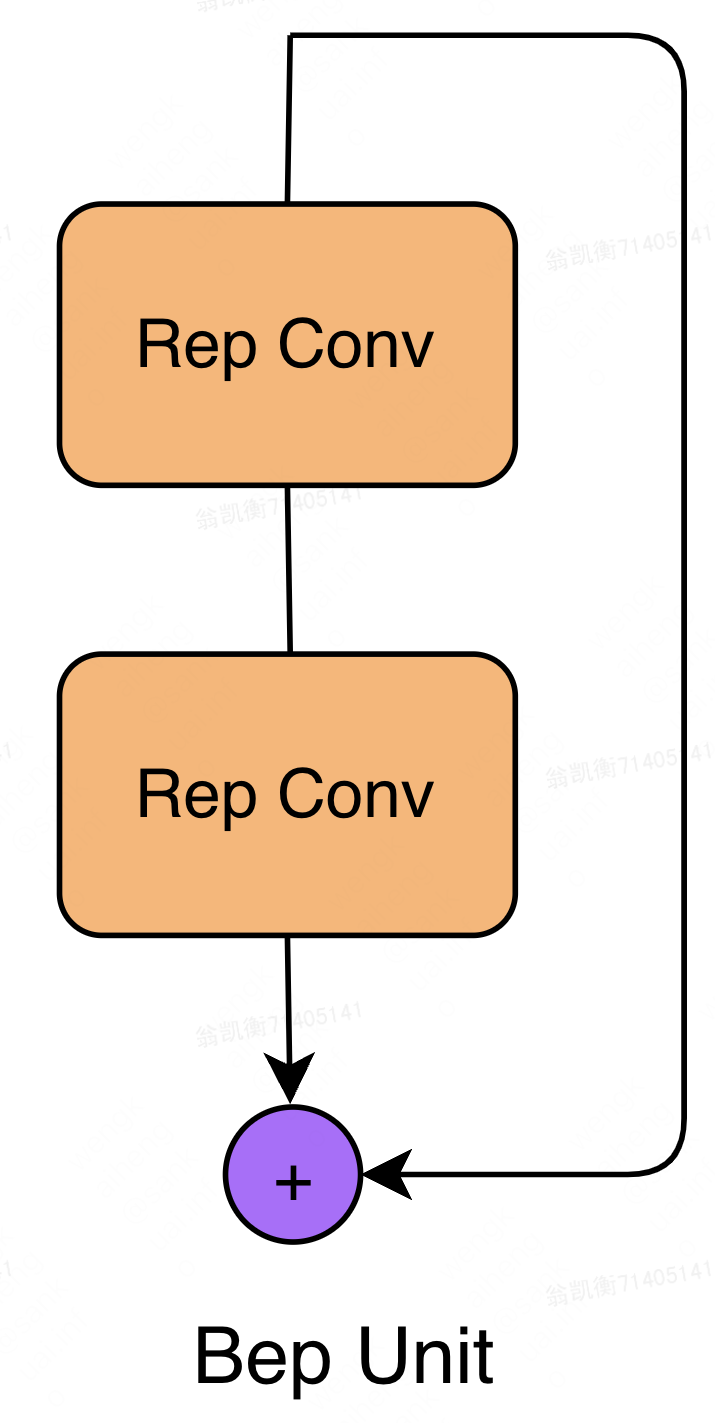}}
\caption{Design of Bep Unit.}
\label{Design of Bep Unit}
\end{figure}

\begin{figure}[t]
\centerline{\includegraphics[width=0.8\linewidth]{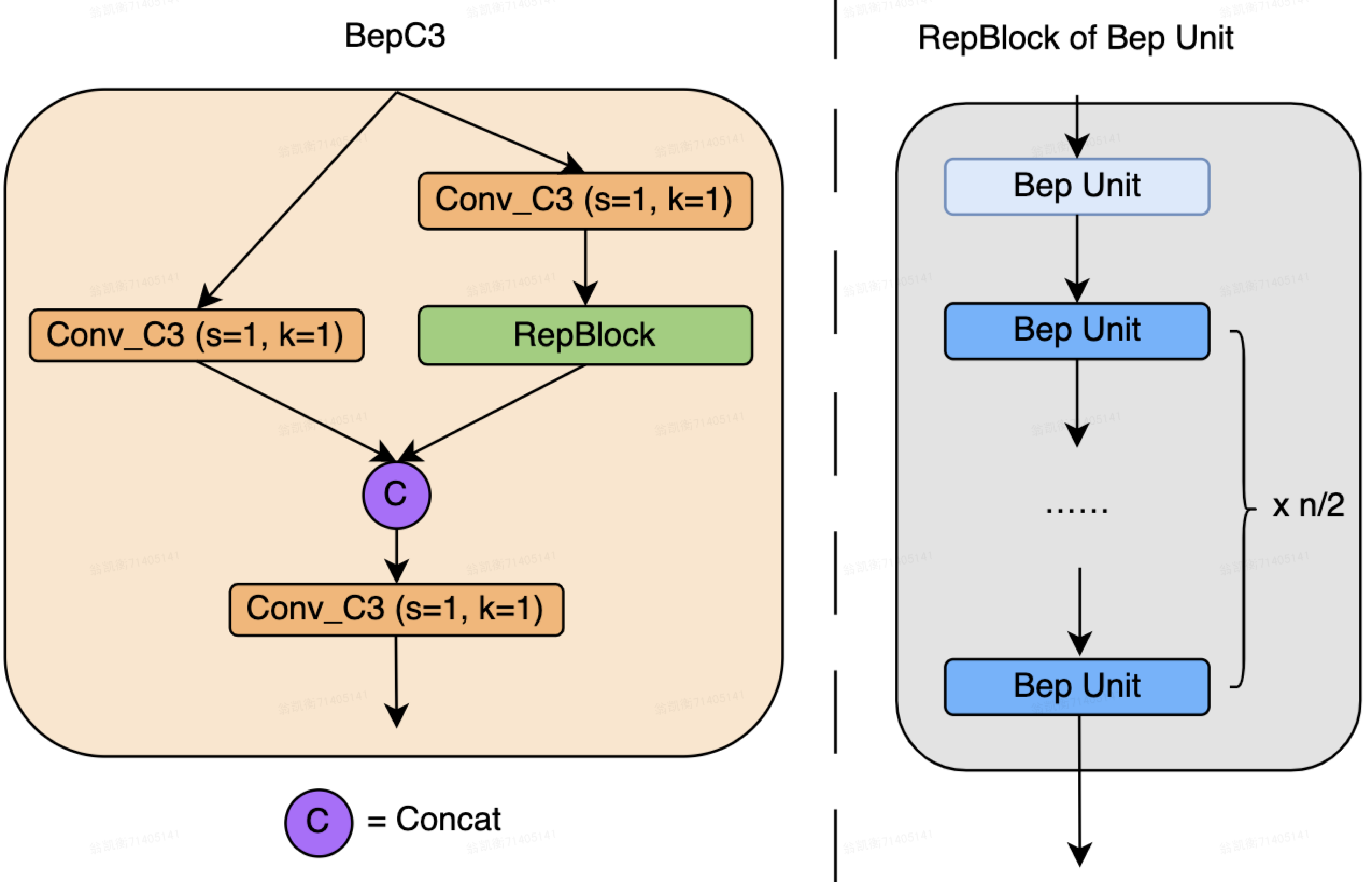}}
\caption{Design of BepC3.}
\label{Design of BepC3}
\end{figure}

However, when YOLOv6-v1 grows to medium size, the inference speed slows down too fast and 
the accuracy is not competitive, compared with csp-style YOLO-series\cite{YOLOv1}\cite{YOLO9000}\cite{YOLOv3}\cite{YOLOv4}\cite{YOLOv5}\cite{PPYOLOE}\cite{YOLOX}. 
As shown in Table \ref{Ablation Study on Structures.}, pure repvgg-style YOLOv6m can not achieve comparable accuracy-speed trade-off. Hence, we explored 
novel structure like multi-path for large size models. 
% to add 

\begin{table}[htbp]
\caption{Ablation Study on Structures.}
\begin{center}
\begin{tabular}{c|c|c|c|c}
\toprule
\textbf{Version} & \textbf{Models} & \textbf{Structures} & \textbf{AP$^{val}$} & \textbf{FPS$_{bs=32}$} \\
\midrule
%\multirow{2}{*}{v1} & YOLOv6-M & Pure Rep-style & 48.8\% & 98 \\
%\midrule to add
%                    & YOLOv6-M & Pure Bep Unit & & \\
%                    & YOLOv6-M & BepC3 Block &  & \\
%\midrule
\multirow{2}{*}{v2} & YOLOv6-M & Pure Rep-style & 47.9\% & 137 \\
                    & YOLOv6-M & BepC3 Block & 48.1\% & 237 \\
\bottomrule
% \multicolumn{4}{l}{$^{\mathrm{a}}$Sample of a Table footnote.}
\end{tabular}
\label{Ablation Study on Structures.}
\end{center}
\end{table}

\subsection{Multi-path Efficient Design}\label{AA}
To solve the problem that pure repvgg-style network can not get accuracy-speed trade-off as expected, we proposed a novel 
structure of Bep unit. The details of Bep unit is shown in Fig \ref{Design of Bep Unit}, numbers of rep convs are connected 
linearly with extra shortcut. With Bep unit, we designed a novel backbone and neck, 
separately named CSPBep and CSPRepPAN. We applied the above structures in YOLOv6-v2 and achieved 
better accuracy-speed trade-off. 

\begin{table}[htbp]
\caption{Specific depth multiplier and width multiplier in YOLOv6.}
\begin{center}
\begin{tabular}{l|c|c}
\hline
%\textbf{Table}&\multicolumn{3}{|c|}{\textbf{Table Column Head}} \\
    & depth multiplier & width multiplier \\
\hline
yolov6n-v1 & 0.33 & 0.25 \\
yolov6s-v1 & 0.33 & 0.50 \\
yolov6n-v2 & 0.33 & 0.25 \\
yolov6s-v2 & 0.33 & 0.50 \\
yolov6m-v2 & 0.60 & 0.75 \\
yolov6l-v2 & 1.0 & 1.0 \\
\hline
\end{tabular}
\label{Specific depth multiplier and width multiplier in YOLOv6}
\end{center}
\end{table}

\begin{table*}[t]
\caption{Comparision of YOLOv6 and Other Detectors.}
\begin{center}
\begin{tabular}{c|c|c|c|c|c}
\toprule
%\textbf{Table}&\multicolumn{3}{|c|}{\textbf{Table Column Head}} \\
Model & Input Size & AP$^{val}$ & FPS$_{bs=1}$ & FPS$_{bs=32}$ & Latency$_{bs=1}$ \\
\midrule
\midrule
YOLOv5-N\cite{YOLOv5} & 640 & 28.0\% & 602 & 735 & 1.7 ms \\
YOLOv5-S\cite{YOLOv5} & 640 & 37.4\% & 376 & 444 & 2.7 ms \\
YOLOv5-M\cite{YOLOv5} & 640 & 45.4\% & 182 & 209 & 5.5 ms \\
YOLOv5-L\cite{YOLOv5} & 640 & 49.0\% & 113 & 126 & 8.8 ms \\
\midrule
\midrule
YOLOX-Tiny\cite{YOLOX} & 416 & 32.8\% & 717 & 1143 & 1.4 ms \\
YOLOX-S\cite{YOLOX} & 640 & 40.5\% & 333 & 396 & 3.0 ms \\
YOLOX-M\cite{YOLOX} & 640 & 46.9\% & 155 & 179 & 6.4 ms \\
YOLOX-L\cite{YOLOX} & 640 & 49.7\% & 94 & 103 & 10.6 ms \\
\midrule
\midrule
PPYOLOE-S\cite{PPYOLOE} & 640 & 43.1\% & 327 & 419 & 3.1 ms \\
PPYOLOE-M\cite{PPYOLOE} & 640 & 49.0\% & 152 & 189 & 6.6 ms \\
PPYOLOE-L\cite{PPYOLOE} & 640 & 51.4\% & 101 & 127 & 10.1 ms \\
\midrule
\midrule
YOLOv7-Tiny\cite{YOLOv7} & 416 & 33.3\% & 787 & 1196 & 1.3 ms \\
YOLOv7-TIny\cite{YOLOv7} & 640 & 37.4\% & 424 & 519 & 2.4 ms \\
YOLOv7\cite{YOLOv7} & 640 & 51.2\% & 110 & 122 & 9.0 ms \\
\midrule
\midrule
YOLOv6-N & 640 & 35.9\% & 802 & 1234 & 1.2 ms \\
YOLOv6-S & 640 & 43.5\% & 358 & 495 & 2.8 ms \\
YOLOv6-M & 640 & 49.5\% & 179 & 233 & 5.6 ms \\
YOLOv6-L & 640 & 51.7\% & 113 & 149 & 8.8 ms \\
\bottomrule
%\cline{2-4} 
%\textbf{Head} & \textbf{\textit{Table column subhead}}& \textbf{\textit{Subhead}}& \textbf{\textit{Subhead}} \\
%\hline
%copy& More table copy$^{\mathrm{a}}$& &  \\
%\hline
% \multicolumn{4}{l}{$^{\mathrm{a}}$Sample of a Table footnote.}
\end{tabular}
\label{Comparision of YOLOv6 and Other Detectors}
\end{center}
\end{table*}

CSP-style\cite{CSPNet} is an efficient design widely used in YOLO-series framework, like YOLOv5\cite{YOLOv5},
PPYOLOE\cite{PPYOLOE} and so on. CSP-style structure uses Cross Stage Partial Network which achieves a richer 
gradient combination while reducing the amount of computation. 
We combined Bep unit with CSP-style structure to design a novel structure named BepC3 
block to balance accuracy and inference speed. The design of BepC3 is described in 
Fig \ref{Design of BepC3}, which is composed of CSP-style structure and Repblock of Bep units. 
As Table \ref{Ablation Study on Structures.} shows the improvements of BepC3 block, the accuracy and speed is  % add 
balanced as expected.

Based on BepC3 block, we separately designed CSPBep backbone and CSPRepPAN neck 
which result in the YOLOv6-v2 models. For CSP-style network in YOLO-series, the 
partial ratio is 1/2 by default. In our design for YOLOv6-v2, we applied partial ratio 
of 2/3 for YOLOv6m and 1/2 for YOLOv6l, aiming to get better performance. 

%\cline{2-4} 
%\textbf{Head} & \textbf{\textit{Table column subhead}}& \textbf{\textit{Subhead}}& \textbf{\textit{Subhead}} \\
%\hline
%copy& More table copy$^{\mathrm{a}}$& &  \\
%\hline
% \multicolumn{4}{l}{$^{\mathrm{a}}$Sample of a Table footnote.}    

\subsection{Scaling Strategy}
Following with YOLOv5, we use scale strategy of depth multiplier and width multiplier to generate various size of models. 
In YOLOv6-v1 and YOLOv6-v2, the depth settings of basic backbones are both [1, 6, 12, 18, 6]. Besides, the width settings are 
[64, 128, 256, 512, 1024]. The depth setting of basic necks are both [12, 12, 12, 12] and the width settings are 
[256, 128, 128, 256, 256, 512]. Table \ref{Specific depth multiplier and width multiplier in YOLOv6} shows the specific depth multiplier and width multiplier applied in 
YOLOv6. 

\section{Experimental Results}
In this section, we show experimental results and details. For object detection, 
experiments are trained on MS COCO-2017 training set with 80 classes and 118k images. 
We use standard COCO AP metric on MS COCO-2017 validation set with 5000 images. 

\subsection{Experiments Details for Object Detection}
For our network design applied in YOLOv6, 
we train the models for a total of 300 epochs with 3 epochs warmup on COCO
train2017. Our training strategy is stochastic gradient descent (SGD) for training 
and an initial lr of 0.01 with the cosine lr scheduler. In our setting, the weight decay is 5e-4. 
For 8-GPU device, the batch size is 256 by default. We also adopt Mosaic and Mixup 
data augmentations and exponential moving average (EMA) during training. 

\subsection{Comparision with Other Detectors}
To present the effects of our network design, Table \ref{Comparision of YOLOv6 and Other Detectors} shows the performance of YOLOv6 models, 
compared with other state-of-the-art object detectors on MS-COCO test split. 
With our optimized models and other improvements, YOLOv6-N/S/M/L models present better accuracy-speed 
trade-off. We evaluate inference speed on NVIDIA Tesla T4 GPU with TensorRT version 7.2, with 
FP16 precision. 

% \subsection{Ablation Study}
% To 

% add some extra experiments
% about 1-2 days.

\section{Conclusion}
In this report, we present our optimization for neural networks applied in YOLOv6. We designed 
Bep unit, BepC3-block/Repblock, EfficientRep/CspBep backbone and Rep-PAN/CSPRepPAN neck. Joined with other improvements, 
we achieved YOLOv6-N/S/M/L models better than other object detectors. 
Meanwhile, we proposed a hybrid network-design strategy applied in various size of models, aiming to 
achieve better accuracy-speed trade-off. 
Moreover, we proposed a novel hardware neural network design with computing and memory balance, which is applied 
in YOLOv6 framework development. We hope that the above proposals can provide inspirations for 
develops and researchers.

% For papers published in translation journals, please give the English 
% citation first, followed by the original foreign-language citation
%  \cite{repvgg}\cite{CSPNet}
%  \cite{YOLOv6}\cite{YOLOX}
%  \cite{MobileOne}\cite{YOLOv5}\cite{TRT-VIT}
%  \cite{YOLOv4}\cite{YOLOv3}
%  \cite{PPYOLOE}\cite{Resnet}.
\bibliographystyle{ieeetr}
\bibliography{references}

\begin{thebibliography}{10}

\bibitem{Inception}
C.~Szegedy, V.~Vanhoucke, S.~Ioffe, J.~Shlens, and Z.~Wojna, ``Rethinking the
  inception architecture for computer vision,'' in {\em 2016 {IEEE} Conference
  on Computer Vision and Pattern Recognition, {CVPR} 2016, Las Vegas, NV, USA,
  June 27-30, 2016}, pp.~2818--2826, {IEEE} Computer Society, 2016.

\bibitem{Resnet}
K.~He, X.~Zhang, S.~Ren, and J.~Sun, ``Deep residual learning for image
  recognition,'' in {\em 2016 {IEEE} Conference on Computer Vision and Pattern
  Recognition, {CVPR} 2016, Las Vegas, NV, USA, June 27-30, 2016},
  pp.~770--778, {IEEE} Computer Society, 2016.

\bibitem{Nasnet}
B.~Zoph, V.~Vasudevan, J.~Shlens, and Q.~V. Le, ``Learning transferable
  architectures for scalable image recognition,'' {\em CoRR},
  vol.~abs/1707.07012, 2017.

\bibitem{AmoebaNet}
E.~Real, A.~Aggarwal, Y.~Huang, and Q.~V. Le, ``Regularized evolution for image
  classifier architecture search,'' {\em CoRR}, vol.~abs/1802.01548, 2018.

\bibitem{MobileOne}
P.~K.~A. Vasu, J.~Gabriel, J.~Zhu, O.~Tuzel, and A.~Ranjan, ``An improved one
  millisecond mobile backbone,'' {\em CoRR}, vol.~abs/2206.04040, 2022.

\bibitem{TRT-VIT}
X.~Xia, J.~Li, J.~Wu, X.~Wang, X.~Xiao, M.~Zheng, and R.~Wang, ``Trt-vit:
  Tensorrt-oriented vision transformer,'' {\em CoRR}, vol.~abs/2205.09579,
  2022.

\bibitem{repvgg}
X.~Ding, X.~Zhang, N.~Ma, J.~Han, G.~Ding, and J.~Sun, ``Repvgg: Making
  vgg-style convnets great again,'' in {\em {IEEE} Conference on Computer
  Vision and Pattern Recognition, {CVPR} 2021, virtual, June 19-25, 2021},
  pp.~13733--13742, Computer Vision Foundation / {IEEE}, 2021.

\bibitem{VGG}
K.~Simonyan and A.~Zisserman, ``Very deep convolutional networks for
  large-scale image recognition,'' in {\em 3rd International Conference on
  Learning Representations, {ICLR} 2015, San Diego, CA, USA, May 7-9, 2015,
  Conference Track Proceedings} (Y.~Bengio and Y.~LeCun, eds.), 2015.

\bibitem{GoogLeNet}
C.~Szegedy, W.~Liu, Y.~Jia, P.~Sermanet, S.~E. Reed, D.~Anguelov, D.~Erhan,
  V.~Vanhoucke, and A.~Rabinovich, ``Going deeper with convolutions,'' in {\em
  {IEEE} Conference on Computer Vision and Pattern Recognition, {CVPR} 2015,
  Boston, MA, USA, June 7-12, 2015}, pp.~1--9, {IEEE} Computer Society, 2015.

\bibitem{one-for-all}
H.~Cai, C.~Gan, T.~Wang, Z.~Zhang, and S.~Han, ``Once-for-all: Train one
  network and specialize it for efficient deployment,'' in {\em 8th
  International Conference on Learning Representations, {ICLR} 2020, Addis
  Ababa, Ethiopia, April 26-30, 2020}, OpenReview.net, 2020.

\bibitem{YOLOv6}
C.~Li, L.~Li, H.~Jiang, K.~Weng, Y.~Geng, L.~Li, Z.~Ke, Q.~Li, M.~Cheng,
  W.~Nie, Y.~Li, B.~Zhang, Y.~Liang, L.~Zhou, X.~Xu, X.~Chu, X.~Wei, and
  X.~Wei, ``Yolov6: {A} single-stage object detection framework for industrial
  applications,'' {\em CoRR}, vol.~abs/2209.02976, 2022.

\bibitem{YOLOv1}
J.~Redmon, S.~K. Divvala, R.~B. Girshick, and A.~Farhadi, ``You only look once:
  Unified, real-time object detection,'' in {\em 2016 {IEEE} Conference on
  Computer Vision and Pattern Recognition, {CVPR} 2016, Las Vegas, NV, USA,
  June 27-30, 2016}, pp.~779--788, {IEEE} Computer Society, 2016.

\bibitem{YOLO9000}
J.~Redmon and A.~Farhadi, ``{YOLO9000:} better, faster, stronger,'' in {\em
  2017 {IEEE} Conference on Computer Vision and Pattern Recognition, {CVPR}
  2017, Honolulu, HI, USA, July 21-26, 2017}, pp.~6517--6525, {IEEE} Computer
  Society, 2017.

\bibitem{YOLOv3}
J.~Redmon and A.~Farhadi, ``Yolov3: An incremental improvement,'' {\em CoRR},
  vol.~abs/1804.02767, 2018.

\bibitem{YOLOv4}
A.~Bochkovskiy, C.~Wang, and H.~M. Liao, ``Yolov4: Optimal speed and accuracy
  of object detection,'' {\em CoRR}, vol.~abs/2004.10934, 2020.

\bibitem{YOLOv5}
glenn jocher~et al, ``yolov5,'' 2021.

\bibitem{PPYOLOE}
S.~Xu, X.~Wang, W.~Lv, Q.~Chang, C.~Cui, K.~Deng, G.~Wang, Q.~Dang, S.~Wei,
  Y.~Du, and B.~Lai, ``{PP-YOLOE:} an evolved version of {YOLO},'' {\em CoRR},
  vol.~abs/2203.16250, 2022.

\bibitem{YOLOX}
Z.~Ge, S.~Liu, F.~Wang, Z.~Li, and J.~Sun, ``{YOLOX:} exceeding {YOLO} series
  in 2021,'' {\em CoRR}, vol.~abs/2107.08430, 2021.

\bibitem{YOLOv7}
C.~Wang, A.~Bochkovskiy, and H.~M. Liao, ``Yolov7: Trainable bag-of-freebies
  sets new state-of-the-art for real-time object detectors,'' {\em CoRR},
  vol.~abs/2207.02696, 2022.

\bibitem{CSPNet}
C.~Wang, H.~M. Liao, Y.~Wu, P.~Chen, J.~Hsieh, and I.~Yeh, ``Cspnet: {A} new
  backbone that can enhance learning capability of {CNN},'' in {\em 2020
  {IEEE/CVF} Conference on Computer Vision and Pattern Recognition, {CVPR}
  Workshops 2020, Seattle, WA, USA, June 14-19, 2020}, pp.~1571--1580, Computer
  Vision Foundation / {IEEE}, 2020.

\end{thebibliography}

\end{document}